\documentclass[fleqn,10pt]{wlscirep}

\usepackage{amsfonts}       % blackboard math symbols
\usepackage{amsmath}
\usepackage{graphicx}
\usepackage{subcaption}
\usepackage{multirow}
\newcommand{\bs}{\boldsymbol}

\newcommand{\ti}{\textit}
\newcommand{\mb}{\mathbf}

\title{Predicting Discharge Medications At Admission Time Based On Deep Learning}

\author[1]{Yuan Yang}
\author[1]{Pengtao Xie}
\author[1]{Xin Gao}
\author[1,3]{Carol Cheng}
\author[1]{Christy Li}
\author[1]{Hongbao Zhang}
\author[1,*]{Eric Xing}

\affil[1]{Petuum Inc., Pittsburgh, 15222, USA}
\affil[3]{Department of Psychiatry, University of Pittsburgh Medical Center, Pittsburgh, 15213, USA}
\affil[*]{eric.xing@petuum.com}

\begin{abstract}
Predicting discharge medications right after a patient being admitted is an important clinical decision, which 
provides physicians with guidance on what type of medication regimen to plan for and what possible changes on initial medication may occur during an inpatient stay. It also facilitates medication reconciliation process with easy detection of medication discrepancy at discharge time to improve patient safety. However, since the information available upon admission is limited and patients' condition may evolve during an inpatient stay, these predictions could be a difficult decision for physicians to make. In this work, we investigate how to leverage deep learning technologies to assist physicians in predicting discharge medications based on information documented in the admission note. We build a convolutional neural network which takes an admission note as input and predicts the medications placed on the patient at discharge time. Our method is able to distill semantic patterns from unstructured and noisy texts, and is capable of capturing the pharmacological correlations among medications. We evaluate our method on 25K patient visits and compare with 4 strong baselines. Our methods demonstrate a 20\% increase in macro-averaged F1 score than the best baseline.
\end{abstract}

\begin{document}

\flushbottom
\maketitle
\thispagestyle{empty}

%\noindent Please note: Abbreviations should be introduced at the first mention in the main text – no abbreviations lists. Suggested structure of main text (not enforced) is provided below.

% ================================================
%                     Introduction
% ================================================
\section*{Introduction}

With the rapid development of new sources of healthcare data extraction, including the adoption of the Electronic Health Record (EHR) systems in the US, a tsunami of medical and healthcare data has emerged~\cite{raghupathi2014big}. As the amount and complexity of the data grows, medical analysis and decision making become time-consuming, error-prone, and suboptimal. One of the most critical and challenging tasks in healthcare is deciding on the treatment plan. Since these are not one-off rulings, but rather a series of decisions made over the course of a patient's journey. For example, a chronic kidney disease elderly patient with chronic heart failure and hypertension could be admitted for a heart failure exacerbation and then require changes to their antihypertnesive medications, such as the adjustments on the type or dosage of diuretics. In cases like this one, it can be difficult for physicians to make informed decision about a trade-off, since, under certain circumstances, a medication may improve one symptoms while worsening another. It would be helpful for physicians to understand which medications should be adjusted through analysis of past cases. Advanced machine learning and deep learning techniques can efficiently digest and leverage information from millions of past admission notes to predict discharge medications with high accuracy. 

In addition, providing predicted discharge medications at admission time will allow doctors to easily detect the discrepancies that may occur at discharge time. Medication discrepancies - unintended differences in documented medication regimens upon admission, transfer, or discharge - affect 70\% of patients and around one third of those discrepancies have the potential to cause moderate to severe harm ~\cite{mueller2012hospital}. In order to identify and resolve discrepancies, it’s crucial to verify and compare patients’ admission and discharge medications (a process called medication reconciliation). Predicting discharge medication at admission time can serve as early-warning tools so that doctors can proactively monitor medications to be prescribed and reduce unnecessary medication errors such as duplications or omissions to ensure patient safety.

Two things make the task of tackling discharge medication predictions more challenging.  
%In this work, we use DL methods to predict discharge medications at admission time. Two issues make this task challenging. 
First, information available upon admission is mostly documented in unstructured clinical notes (called admission notes), covering past medical history, family history, allergies, etc. Compared to structured information like labs and vital signs, these free-form texts with synonyms, abbreviations and misspellings are difficult for machines to process and understand. Distilling semantic patterns from such unstructured and noisy texts is very challenging. 
%Second, since independent medications have pharmacological effects on each other, there are various situations in medicine where medications are prescribed together as a pair or more due to research and public consensus about optimal therapy. For example, for those patients who have had a recent stroke while already on aspirin, it is recommended as future stroke prevention for them to be prescribed aspirin and clopidogrel, which is referred to as dual anti-platelet therapy. In another example, in those who have had a recent heart attack while also having diabetes, it is recommended starting them on a statin, aspirin, beta-blocker, and ace-inhibitor. In this situation, physicians utilize multi-medication therapy because all have been shown to have a certain impact on mortality/disease progression. How to automatically discover and leverage such pharmacological correlations among medications is crucial for more accurate multiple-medication prediction and is highly non-trivial. 
Second, in order to quickly and effectively cure diseases and meet treatment goals, two or more than two medications are commonly prescribed since combination medications are widely adopted by the clinical guidelines or medical consensus. For example, patients who have experienced a recent stroke while already on aspirin are recommended dual antiplatelet therapy with aspirin and clopidogrel for future stroke prevention. Physicians tend to utilize multi-medication therapy because it has been shown to have an impact on mortality/disease progression. Automatically discovering and leveraging such pharmacological correlations among medications is crucial for more accurate multiple-medication prediction and is highly non-trivial. 

\section*{Contributions}
In this paper, we develop a Convolutional Neural Network (CNN) model which takes an admission note as input and predicts one or multiple discharge medications. The CNN model is able to learn rich semantic representations from raw texts and can automatically capture the correlations among medications. We evaluate the model on a specific medication category -- antihypertensive medications, on 25K patient visits. Our model demonstrates a 20\% increase in macro-averaged F1 score than the best baseline method. 
%In addition, the predictions made by our model is interpretable.

% ================================================
%                      Related work
% ================================================
\section*{Related work}

%In the past few decades, clinical decision support systems (CDSSs) have been widely adopted to improve health care~\cite{osheroff2007roadmap} quality, which takes advantage of pre-compiled domain knowledge to address clinical needs such as ensuring accurate diagnoses, screening for preventable diseases, or averting adverse drug events~\cite{garg2005effects}. However, the results of the research to date are mixed in terms of the effectiveness of CDSS for particular conditions or particular types of CDSS~\cite{berner2009clinical}. And the attempts of incorporating EHR in CDSS usually seems to be fruitless~\cite{moja2014effectiveness}. 
With the prosperity of healthcare data such as Electronic Health Record (EHR), genomic data, patient behavior data and the growing need of extracting knowledge and insights from these data, data-driven healthcare analytics based on machine learning and deep learning have received much attention recently. % methods have been widely applied to

\paragraph*{Predictive Modeling}

Predictive modeling in data-driven healthcare is concerned about building machine learning models to predict diagnosis, prognosis, patient risk factors, readmission, disease onset and so on. Wang et al.~\cite{wang2012medical} studied disease prognosis by leveraging the information of clinically
similar patient cohort. Zhou et al.~\cite{zhou2013patient} proposed a top-k stability selection method to select the most informative features for patient risk prediction. Chen et al.~\cite{chen2015cloud} developed a cloud-based predictive modeling system for pediatric asthma readmission prediction. Lipton et al.~\cite{lipton2015learning} applied long short-term memory (LSTM) network for pattern discovery from lab measurements and make diagnostic predictions based on the discovered patterns. Razavian et al.~\cite{razavian2015population} developed predictive models to predict the onset and risk factors of type-2 diabetes. Choi et al.~\cite{choi2016doctor} used recurrent neural networks (RNN) to predict diseases and recommend medications for patients' subsequent encounters~\cite{miotto2017deep}. Pham et al.~\cite{pham2016deepcare} utilized RNN to model disease progression, recommend intervention and predict future risk for mental health and diabetes. Razavian et al.~\cite{razavian2016multi} developed a LSTM network and two convolutional neural networks (CNNs) for multi-task prediction of disease onset. Several works~\cite{cheng2016risk,zhou2014micro,nguyen2017mathtt} leveraged CNNs for heart disease diagnosis. Miotto et al.~\cite{miotto2016deep} proposed to use denoising autoencoders to learn a general-purpose patient representation that can be used to predict patients' future health status.

\paragraph*{Natural Language Processing and Understanding of Clinical Notes}
Clinical notes contain rich medical information. Many studies have developed natural language processing and machine learning methods to extract useful information from free-from clinical texts. Cheng et al.~\cite{chen2013applying} investigated word sense disambiguation using support vector machine (SVM) and active learning.
Fan et al.~\cite{fan2013syntactic} developed a handbook of domain-customized syntactic parsing guidelines of clinical texts, in particular, how to handle ill-formed sentences. Tang et al.~\cite{tang2013hybrid} developed a temporal information extraction system that can identify events, temporal expressions, and their temporal relations in clinical text.
Tang et al.~\cite{tang2013recognizing} studied clinical entities recognition in hospital discharge summaries using structured SVM. Gobbel et al.~\cite{gobbel2014assisted} designed a tool to assist the annotation of medical texts based on interactive training. Lei et al.~\cite{lei2014comprehensive} systematically investigated features and machine learning algorithms for named entity recognition in Chinese clinical text. Tang et al.~\cite{tang2014evaluating} performed a comparison study of three different types of word representation features for biomedical named entity recognition, including clustering-based representation, distributional representation, and word embeddings. Halpern et al.~\cite{halpern2016clinical} developed a bipartite probabilistic graphical models for joint prediction of clinical conditions from the electronic medical records. Jagannatha et al.~\cite{jagannatha2016structured} applied LSTM and conditional random field to recognize medical entities from clinical notes.

% ================================================
%                        Methods
% ================================================
\section*{Methods}

\begin{figure}[tb]
\centering
\includegraphics[width=0.6\columnwidth, trim={0cm 0 0cm 0cm}, clip]{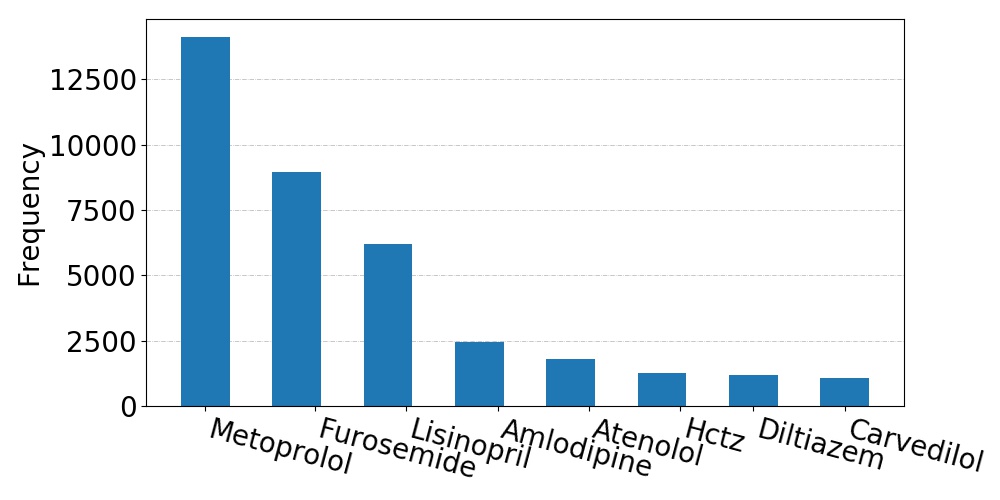}
\caption{Frequencies of 8 antihypertensive medications in the MIMIC-III dataset. ``Hctz'' is the shorthand for hydrochlorothiazide.}
\label{fig:medfreq}
\end{figure}
%We evaluate the performance of CNN model in predicting the discharge medication on a public dataset containing past admission records and individual EHRs. 

%The input is assumed to be unstructured text, and the prediction is compared to the ground-truth medication list for measuring the performance.

\subsection*{Study design} On previously collected electronic health records of Intensive Care Unit (ICU) patients, we performed a retrospective study to build a deep learning model predicting discharge medications based on admission notes. We evaluated the proposed model by comparing the predicted medications with the ground-truth medications given by physicians and compared with several competitive baselines.

\subsection*{Dataset description and preparation} The study is performed on the MIMIC-III dataset~\cite{johnson2016mimic}, a public dataset containing 52K patient-visits of 40K unique patients who stayed in the ICU at Beth Israel Deaconess Medical Center between 2001 and 2012 (registration required for accessing the data). This dataset contains thousands of unique medications. With such limited data (52K patients visits), it is extremely difficult to build a highly accurate predictive model for thousands of medications. Instead, we focus our study on 8 antihypertensive medications: metoprolol, furosemide, lisinopril, amlodipine, atenolol, hydrochlorothiazide (hctz), diltiazem, carvedilol, due to the following two considerations. First, these medications are commonly used to treat hypertension -- one of the most prevalent and severe chronic diseases. In particular, they widely occur in MIMIC-III. Second, they are difficult to predict: their prescriptions and usage have large variance, especially when multiple of them are used together, making them an excellent testbed of our approach. The frequencies of these medications are imbalanced (as shown in Figure~\ref{fig:medfreq}): some of them occur very frequently while others are of small frequency, which adds another layer of difficulty for prediction. It is worth noting that our method can be readily extended to other categories of medications, as long as sufficient clinical data is available. 

%\yy{as a comorbidity with acute diseases, since patient condition can get very complicated, planing personalized hypertension treatment while not exaggerating the acute meds are difficult and lacks guidance. This can expose how model performs whiling considering the patient's health status as a whole, make it tries to balance, come up with the an optimal plan.}. 

In this dataset, only discharge notes and nursing notes are provided. The admission notes are not directly accessible. To address this issue, from the discharge notes which contain information collected during the entire course of a patient visit, we extract the information that is only available upon admission such as chief complaint, past medical history and combine them into an admission note. The entire list of different types of admission information is given in Table \ref{tab:syns}. Each discharge note consists of multiple sections, each of which has a heading summarizing the content of this section. We perform string matching on the headings to pick up the sections containing admission information. It is often the case that the same type of admission information (e.g., history of the present illness) are given different headings (e.g., ``History of present illness'', ``HPI'') in different notes. To extract admission information as much as possible, we enumerate all the heading strings that indicate admission information (example strings shown in Table~\ref{tab:syns}) and use them to match the headings in the notes to pick out sections containing admission information.

The discharge medications are also contained in the admission notes as semi-structured sections. Each such section starts with a heading named ``discharge medications'' or ``meds on discharge", which is followed by a list of bullet points, each containing a discharge-medication string. We use regular expressions to recognize the antihypertensive medications. Patient visits containing no antihypertensive medications are discarded. 

After preprocessing, the dataset contains 25K patient visits, each consisting of an admission note and a set of discharge antihypertensive medications. Words in the admission notes are all lower-cased and stopwords are removed. The average word count of admission notes is 350.

\begin{table}[t]
\centering
\begin{tabular}{l|l}
\hline
Types of Admission Information              & Heading Strings                                                                              \\ \hline
Allergy                     & “allergies”                                                                                    \\
Chief complaint            & “chief complaint”                                                                              \\
History of present illness & “history of present illness”, “hpi”                                                            \\
Past medical history      & “past medical history”, “major surgical or invasive procedure”, 
           \\
Social history             & “social history”, “history”                                                                    \\
Family history             & “family history”, “family hx”                                                                  \\
Initial exam               & “admission labs”, “physical exam”                                                              \\
%Discharge medications      & “discharge medications”, “meds on discharge”\\
Admission medications      & “admission medications”, “meds on admission” \\ \hline
\end{tabular}
\caption{Different types of information available upon admission and subsets of heading strings used to extract each type of information.}

%For each type, one or more synonym strings are used to
%Heading groups and corresponding subsets of synonyms: when processing each note, paragraphs with its heading belonging to any of the group listed below are retained and grouped with other contents within the same group, while others are removed.}
\label{tab:syns}
\end{table}

\begin{figure}[tb]
\centering
\includegraphics[width=.8\columnwidth, trim={5cm 1.5cm 3.5cm 1.5cm}, clip]{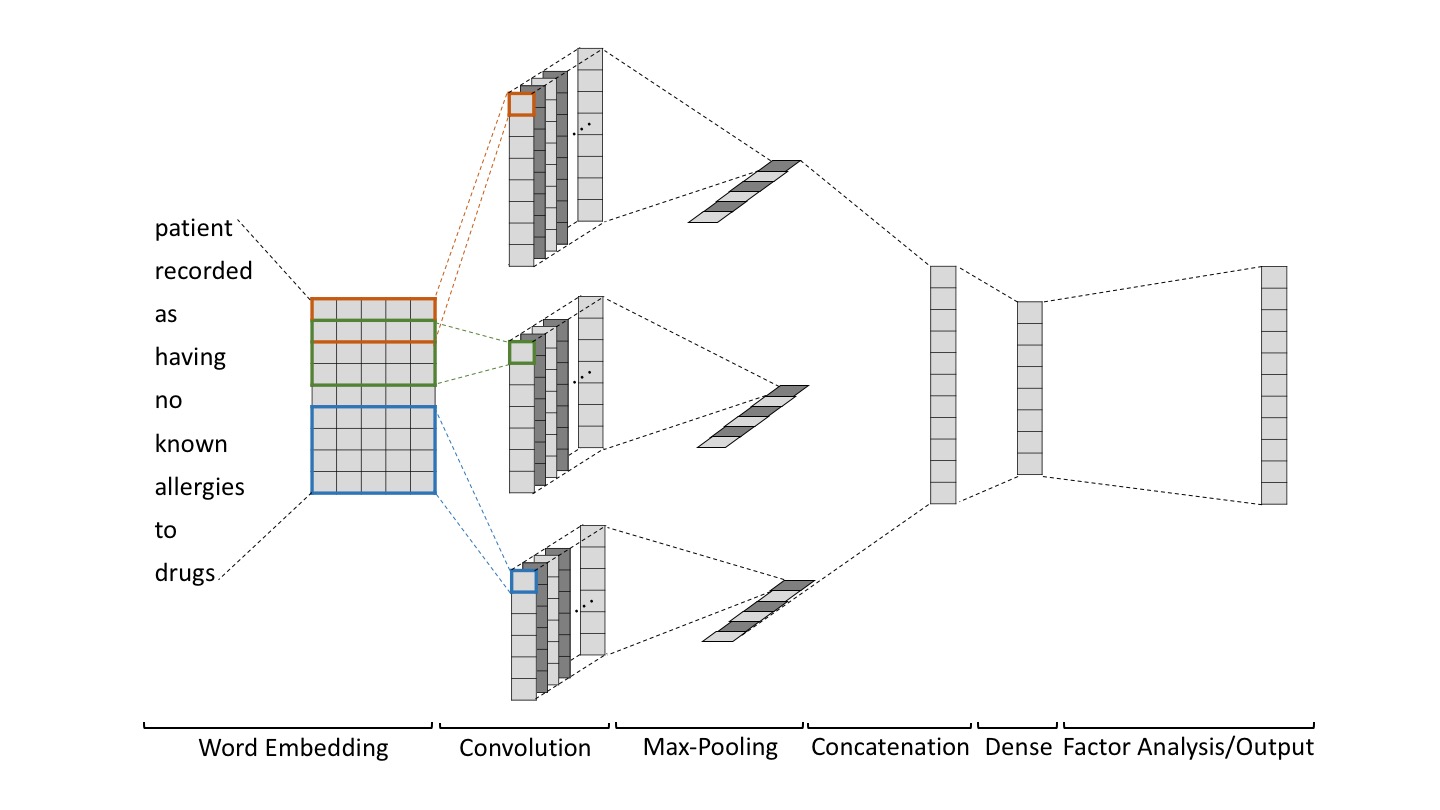}
\caption{The network architecture of the CNN model. 
%The overall architecture of CNN with 3 channels of convolution layers with window sizes of 2,3 and 4. Diagram adopted from Kim et al.~\cite{kim2014convolutional}.
}
\label{fig:arch}
\end{figure}

\subsection*{Model design}
We develop a deep learning model to predict discharge medications based on information available upon admission. The input of the model is an admission note and the output is one or multiple medications that are placed on the patient at discharge time. The model is supposed to have two capabilities. First, it should be able to effectively distill high-level semantics from the noisy and unstructured raw texts and properly take into account the sequential structure among consecutive words. A sequence of $n$ consecutive words is called a $n$-gram where $n$ is the size of this gram. Second, it should have the mechanism to capture the pharmaceutical correlation among medications. 

To simultaneously achieve these two goals, we develop a convolutional neural network (CNN)~\cite{kim2014convolutional} based model. At a high level, the model (1) uses multiple layers of stacked hidden units to capture the latent semantics of input notes; (2) uses convolutional operators with different window sizes to capture the local semantics and sequential structure existing in $n$-grams; (3) discovers common latent factors to capture the pharmaceutical correlations among medications. In the sequel, we describe the model in detail.

The full architecture of the CNN model is shown in Figure~\ref{fig:arch}. The input of the model is an admission note, which is represented as a sequence of $l$ words $t_1, t_2, \cdots, t_l$, where $t_i$ denotes the $i$-th word. $l$ is the length of this note. 
%tunable hyper-parameter and is fixed for notes of any length. If a note has more than $l$ words, only the first $l$ words are retained and the rest are discarded. On the other hand, if a note has less than $l$ words, it is repetitively padded with a special word ``PADDING'' until the number of words reaches $l$. 
The first layer of the CNN model performs \textit{word embedding}, which learns a semantic vector representation for each word. Different occurrences of the same word (e.g., $t_1$ and $t_8$ are both ``hypertension'') share the same embedding vector. This layer has a weight parameter matrix $\mb{T}\in\mathbb{R}^{v\times h}$ where $v$ is the number of unique words in the vocabulary, $h$ is the dimension of the embedding vectors and the $i$-th row vector is the embedding of word $i$. After the word embedding layer, the note is represented as a $l$-by-$h$ matrix $\mb{D}$ where the $i$-th row vector is the embedding of $t_i$. 

The second layer of CNN performs \ti{convolution}, which captures the local semantics and sequential structure manifested in $n$-grams. The convolution operation is conducted by \ti{filters}, each aiming at capturing one specific semantic. Each filter is parameterized by a $n$-by-$h$ weight matrix $\mb{W}$ and a bias term $b$, where $n$ is the window size. The input of the filter is a $n$-by-$h$ matrix $\mb{D}_{i:i+n-1}$ --- the embeddings of words $t_i, \cdots, t_{i+n-1}$ in a $n$-gram. The output is a new feature value computed as $$c_i=f(\langle \mb{W}, \mb{D}_{i:i+n-1}\rangle +b),$$ where $\langle\cdot, \cdot, \rangle$ denotes matrix inner product and $f(\cdot)$ denotes a nonlinear activation function such as sigmoid, hyperbolic tangent and rectified linear. These nonlinear functions enhance the expressiveness of CNN to capture nonlinear and complicated semantic patterns. The same filter is applied to all possible $n$-grams at different positions $i$ to detect the saliency of the semantic it represents. To capture a variety of different semantics, multiple filters with different weight parameters are used. In addition, the size of grams plays a crucial role in differentiating the subtlety in semantics. For example, for the three grams ``leukemia'', ``lymphoblastic leukemia'', ``acute lymphoblastic leukemia'' with size 1, 2, 3 respectively, though they all refer to the same type of disease, the specificity of disease type varies substantially under different sizes. To be able to capture semantically meaningful phrases containing a variable number of words, we set up filters with different window sizes.

The convolution layer is followed by the \ti{max pooling} layer, which aims at capturing the most important semantics in the notes. 
%The idea is as follows: for each filter, we retain the most salient 
Among the feature values $c_1,\cdots,c_{l-n+1}$ produced by a filter, the max pooling scheme selects the largest one as the final output of this filter. The purpose is to be robust to small variations in semantics. The pooled values of all filters are concatenated together to represent the entire note. The concatenation layer is followed by a dense layer, aiming at further capturing the nonlinear patterns presenting in notes. The dense layer consists of $s$ hidden units, each of which is connected to all units (denoted by $\mb{z}$) in the concatenation layer. The value of the $i$-th dense unit is computed as
$$x_i=f(\mb{u}_i^\top\mb{z}+d_i),$$ 
where $\mb{u}_i$ and $d_i$ are the weight vector and bias term associated with this unit, and $f$ is the activation function. 

Next, on top of the dense layer, we make predictions of the discharge medications. As noted earlier, 
%So far, we have obtained the representation learning module for admission notes. Next, on top of these representations, we build a prediction module for discharge medications. 
it is often the case that multiple medications are used together. Thereby, this is a multi-label prediction problem: given a total of $k$ medications, picking up a subset of them. As stated earlier, medications have pharmacological effects on each other and it is important to capture such correlation for more accurate prediction. To achieve this goal, we borrow ideas from the factor analysis~\cite{jolliffe1986principal} model and assume these medications share a common set of latent factors (which are probabilistic variables) and the predictive model of each medication is a linear combination of the latent factors. The linear coefficients are different for different medications. These common factors capture the commonality among medications while the medication-specific linear coefficients preserve the difference among medications. We use the hidden units in the dense layer to characterize the latent factors and model each medication as a weighted combination of the factors (denoted by $\mb{x}$). Let $\mb{y}$ be a $k$-dimensional random vector where $y_i$ denotes the confidence score of using medication $i$. According to the factor analysis model, we assume
$$\mb{y} = \bs\mu + \bs\Lambda\mb{x},$$
where $\bs\Lambda\in\mathbb{R}^{k\times s}$ is the coefficient matrix (referred to as loading matrix in the factor analysis literature) and $\bs\mu\in\mathbb{R}^k$ is an offset vector. The expectation of $\mb{y}$ is:
$$
\mathbb{E}[\mb{y}]=\mathbb{E}[\bs\mu + \bs\Lambda\mb{x}]=\bs\mu +\bs\Lambda \mathbb{E}[\mb{x}].
$$
The covariance matrix of $\mb{y}$ is calculated as:
\begin{equation}
\label{eq:covy}
\begin{array}{lll}
\hspace{3.2cm}
\textrm{cov}[\mb{y}] & = & \mathbb{E}[(\mb{y}-\mathbb{E}[\mb{y}])(\mb{y}-\mathbb{E}[\mb{y}])^\top] \\
& = & \mathbb{E}[(\bs\mu + \bs\Lambda\mb{x}-(\bs\mu +\bs\Lambda \mathbb{E}[\mb{x}]))(\bs\mu + \bs\Lambda\mb{x}-(\bs\mu +\bs\Lambda \mathbb{E}[\mb{x}]))^\top] \\
& = & \bs\Lambda\mathbb{E}[(\mb{x}-\mathbb{E}[\mb{x}])(\mb{x}-\mathbb{E}[\mb{x}])^\top]\bs\Lambda^\top \\
& = & \bs\Lambda\textrm{cov}[\mb{x}]\bs\Lambda^\top \\
\end{array}
\end{equation}
This covariance matrix, which captures the correlations among medications, is automatically learned during the training process.
%This factor analysis model is implicitly learned through the learning process and enables the CNN model to capture the commonality among medications while the medication-specific linear coefficients preserve the difference among medications.
Given these confidence scores, we transform them into probabilities. Specifically, 
%On top of these latent factors, we predict the presence/absence of each medication. 
the probability of using medication $i$ is calculated as
$$p_i=1/(1+\exp(-y_i)).$$

To this end, we obtain the entire CNN architecture which takes an admission note as input and produces the prediction probabilities of discharge medications. Next, we discuss how to learn the model parameters including word embedding vectors, weight vectors and bias terms from physician-labeled training data. Let $\{(\phi^{(j)},\mb{y}^{(j)})\}_{j=1}^{m}$ denote $m$ training examples where $\phi^{(j)}$ is the admission note of the $j$-th example and $\mb{y}^{(j)}\in\mathbb{R}^k$ (where $k$ is the number of unique medications) is a binary vector representing the ground-truth discharge medications labeled by physicians. $y_i^{(j)}=1$ denotes that the $i$-th medication was used. Let $\mb{p}^{(j)}=\Phi(\phi^{(j)})$ be a $k$-dimensional prediction vector where $p_i^{(j)}$ is the probability that medication $i$ is placed on patient $j$ and $\Phi(\cdot)$ is the mapping induced by CNN. 
%denote the prediction probabilities generated by the mapping induced by CNN. 
We define a loss function to measure the discrepancy between the prediction $\mb{p}^{(j)}$ and the ground-truth $\mb{y}^{(j)}$. A commonly used loss function is the cross entropy loss: $$-\sum_{j=1}^{m}\sum^k_{i=1} y_{i}^{(j)} \log p_i^{(j)} + (1-y_{i}^{(j)})\log(1 - p_i^{(j)}).$$ 
%Given $m$ training examples, we define an overall training loss by adding the cross entropy loss of individual examples together, 
We then minimize this loss to obtain the optimal weight parameters of the CNN model using a gradient descent algorithm, which iteratively performs the following two steps until convergence: (1) taking derivatives of the loss function with respect to all weight parameters using a back-propagation procedure; (2) moving each weight parameter by a small step along its negative-gradient direction. In model training, batch normalization~\cite{ioffe2015batch} is performed before every non-linear transform; dropout~\cite{srivastava2014dropout} is performed on the hidden units after concatenation; and L2 regularization is applied on all learnable weight parameters except bias terms. These techniques help to prevent over-fitting and speed up training.

%To this end, we obtain the prediction probabilities $\{p_i\}_{i=1}^K$ of all medications. 

Given the learned model, we use it to make predictions on previously unseen patient visits: (1) calculating the probability of each medication; (2) picking up the medications whose probabilities are larger 0.5 as discharge medications.

\subsection*{Setup and Evaluation}

\textbf{Baseline models}: we compare the CNN model against three classification models, including support vector machine~\cite{cortes1995support} (SVM), random forest~\cite{ho1995random} (RF) and logistic regression~\cite{cox1958regression} (LR). Unlike CNN, these models lack the mechanism to automatically learn semantic representations of texts or capture correlations among medications. The input features of these models are term-frequency and inverse document frequency (TF-IDF) vectors extracted from the admission notes. TF-IDF is widely used in natural language processing and text classification.  

As observed in clinical practice, discharge medications are closely related to admission medications. Sometimes, they are even largely overlapped. One may wonder whether it is sufficient to predict discharge medications solely based on admission medications, without the need of other types of admission information such as past medical history and chief complaint. To answer this question, we compare with another baseline, which only uses admission medications as input and uses a multi-layer perceptron (MLP) to predict the discharge medication. For each note, we extract the admission medications from the section titled ``Admission medications" or ``Meds on admission" and encode them into a $s$-dimensional binary vector $\mb{x}$ where $x_i=1$ denotes that this note has the $i$-th admission medication and $s$ is the number of unique admission-medications. $\mb{x}$ is fed into the MLP for discharge medication prediction. 

\textbf{Experimental setup}: CNN is implemented with TensorFlow~\cite{abadi2016tensorflow} and its weight parameters are optimized using the Adam~\cite{kingma2014adam} method, a variant of the stochastic gradient descent (SGD) algorithm. The dataset is split into 3 parts: 80\% for training, 10\% for validation and 10\% for testing. The hyper-parameters are tuned using grid search on the validation set. In CNN, we set three sets of filters. Filters' window size in each set is 3, 4, 5 respectively. Each set contains 64 filters. The learning rate in Adam is set to 0.01. The keep rate in Dropout~\cite{srivastava2014dropout} is set to 0.3. The L2 regularization parameter is set to 0.1. The TF-IDF feature dimension is set to 1500 for SVM and RF and is set to 2500 for LR. In SVM, we use the radius basis function (RBF) kernel where the scale parameter is set to 0.1. The tradeoff parameter between hinge loss and margin size is set to $2^{15}$. In RF, the number of trees is set to 200; the number of features to consider when looking for the best split is set to 80; the minimum number of samples required to be at a leaf node is set to 1. In LR, the L2 regularization parameter is set to 1.

\textbf{Evaluation metrics}: To evaluate model performance, we measure class(medication)-wise precision, recall and F1 scores on the test set and the micro and macro averages of these scores across all medication classes. The micro average is obtained by taking the weighted average of class-wise scores where the weights are proportional to the frequencies of medication classes. The macro average is calculated as the unweighted average of the class-wise scores. Macro-average gives equal weight to each medication class, whereas micro average gives equal weight to each data example. Since the frequencies of medication classes are highly skewed (as shown in Figure~\ref{fig:medfreq}), micro-average -- which is in favor of frequent classes -- may underestimate the error in infrequent classes. The macro-average score alleviates this drawback by treating each class equally and can better reveal model's performance on infrequent classes.

% ================================================
%                       Results
% ================================================
\section*{Results}

Table \ref{tab:res} shows the precision (P), recall (R) and F1-score (F) of different methods for each antihypertensive medication averaged over 5 runs with random dataset split (standard deviations provided in Supplementary Table \ref{tab:std}). The last two lines show the micro- and macro-averaged scores over all medications. As can be seen, CNN achieves much better micro and macro average F1 than the baseline methods. Between the two averages, CNN's improvement on macro average is more significant. On 7 medications, CNN achieves the best F1 scores. The only exception is furosemide, where RF outperforms CNN. CNN's improvement over the baselines is mainly on the recall scores while its precision scores are comparable with the baselines. Among the baselines, MLP which only uses admission medications as inputs performs the worst in terms of the micro and macro average F1. SVM and RF which are non-linear models perform better than LR which is a linear model.

% \usepackage{multirow}
% \usepackage{graphicx}
\begin{table}[t]
\centering
\resizebox{\textwidth}{!}{%
\begin{tabular}{c|lll|lll|lll|lll|lll}
\hline
\multirow{2}{*}{Medication} & \multicolumn{3}{c|}{CNN}                                               & \multicolumn{3}{c|}{MLP}                                               & \multicolumn{3}{c|}{SVM}                                               & \multicolumn{3}{c|}{RF}                                                & \multicolumn{3}{c}{LR}                                                \\ \cline{2-16} 
                            & \multicolumn{1}{c}{P} & \multicolumn{1}{c}{R} & \multicolumn{1}{c|}{F} & \multicolumn{1}{c}{P} & \multicolumn{1}{c}{R} & \multicolumn{1}{c|}{F} & \multicolumn{1}{c}{P} & \multicolumn{1}{c}{R} & \multicolumn{1}{c|}{F} & \multicolumn{1}{c}{P} & \multicolumn{1}{c}{R} & \multicolumn{1}{c|}{F} & \multicolumn{1}{c}{P} & \multicolumn{1}{c}{R} & \multicolumn{1}{c}{F} \\ \hline
 Metoprolol                  & 0.73                  & 0.87                  & 0.79                  & 0.64                  & 0.96                  & 0.77                  & 0.74                  & 0.79                  & 0.76                  & 0.74                  & 0.89                  & \textbf{0.81}         & 0.74                  & 0.85                  & 0.79                  \\
Furosemide                  & 0.59                  & 0.85                  & \textbf{0.70}         & 0.62                  & 0.42                  & 0.46                  & 0.70                  & 0.64                  & 0.67                  & 0.73                  & 0.67                  & \textbf{0.70}         & 0.74                  & 0.62                  & 0.68                  \\
Lisinopril                  & 0.56                  & 0.51                  & \textbf{0.53}         & 0.63                  & 0.43                  & 0.49                  & 0.56                  & 0.44                  & 0.49                  & 0.70                  & 0.35                  & 0.46                  & 0.66                  & 0.29                  & 0.40                  \\
Amlodipine                  & 0.59                  & 0.45                  & \textbf{0.49}         & 0.11                  & 0.08                  & 0.09                  & 0.61                  & 0.34                  & 0.43                  & 0.72                  & 0.29                  & 0.41                  & 0.78                  & 0.10                  & 0.18                  \\
Atenolol                    & 0.58                  & 0.27                  & \textbf{0.32}         & 0.00                  & 0.00                  & 0.00                  & 0.45                  & 0.23                  & 0.30                  & 0.70                  & 0.14                  & 0.24                  & 0.52                  & 0.03                  & 0.06                  \\
Hctz         & 0.41                  & 0.23                  & \textbf{0.26}         & 0.00                  & 0.00                  & 0.00                  & 0.35                  & 0.21                  & \textbf{0.26}         & 0.61                  & 0.04                  & 0.07                  & 0.60                  & 0.02                  & 0.04                  \\
Diltiazem                   & 0.62                  & 0.37                  & \textbf{0.46}         & 0.00                  & 0.00                  & 0.00                  & 0.45                  & 0.21                  & 0.28                  & 0.73                  & 0.12                  & 0.20                  & 0.85                  & 0.04                  & 0.07                  \\
Carvedilol                  & 0.60                  & 0.56                  & \textbf{0.57}         & 0.00                  & 0.00                  & 0.00                  & 0.51                  & 0.28                  & 0.36                  & 0.74                  & 0.27                  & 0.40                  & 0.79                  & 0.08                  & 0.15                  \\ \hline
Micro Avg                   & 0.63                  & 0.70                  & \textbf{0.65}         & 0.51                  & 0.54                  & 0.51                  & 0.65                  & 0.58                  & 0.60                  & 0.72                  & 0.60                  & 0.61                  & 0.72                  & 0.53                  & 0.55                  \\
Macro Avg                   & 0.59                  & 0.51                  & \textbf{0.52}         & 0.25                  & 0.24                  & 0.23                  & 0.55                  & 0.39                  & 0.44                  & 0.71                  & 0.35                  & 0.41                  & 0.71                  & 0.25                  & 0.30                  \\ \hline

\end{tabular}%
}
\caption{Medication-wise precision (P), recall (R) and F1-score (F) for CNN and 4 baseline models averaged over 5 runs. From top to bottom, medications are shown in descending order of their frequencies. The overall performance is measured using micro- and macro-average.
%Medications are sorted by dataset frequency. Model overall performance is evaluated by micro- and macro-averaging the precision, recall and F1-score over all classes.
}
\label{tab:res}
\end{table}

\begin{table}[t]
\centering
\begin{tabular}{llllllll}
\hline
%\multicolumn{8}{c}{Nearest neighbors of a subset of words}                                                                                                            \\ \hline
Query       & \multicolumn{1}{l|}{NN}         & Query      & \multicolumn{1}{l|}{NN}         & Query        & \multicolumn{1}{l|}{NN}         & Query      & NN              \\ \hline
artery      & \multicolumn{1}{l|}{troponin}   & pressure   & \multicolumn{1}{l|}{quinapril}  & fibrillation & \multicolumn{1}{l|}{diltiazem}  & lopressor  & metoprolol      \\
glucose     & \multicolumn{1}{l|}{blood}      & renal      & \multicolumn{1}{l|}{amlodipine} & sclera       & \multicolumn{1}{l|}{yellow}     & vessel     & triple          \\
aortic      & \multicolumn{1}{l|}{mitral}     & lisinopril & \multicolumn{1}{l|}{pindolol}   & diarrhea     & \multicolumn{1}{l|}{nonbloody}  & angina     & echocardiograms \\
edema       & \multicolumn{1}{l|}{myxedema}   & graft      & \multicolumn{1}{l|}{mqwmi}      & erythema     & \multicolumn{1}{l|}{cellulitis} & amlodipine & norvasc         \\
ventricular & \multicolumn{1}{l|}{aneurismal} & lasix      & \multicolumn{1}{l|}{furosemide} & cardiac      & \multicolumn{1}{l|}{echoes}     & incision   & open  \\ \hline
\end{tabular}
\caption{A subset of (query) words together with their nearest neighbor (NN). The ``nearness'' between words is measured as the Euclidean distance between their corresponding embedding vectors.}
\label{tab:knn}
\end{table}

\begin{figure}[tb]
\centering
\includegraphics[height=.45\columnwidth]{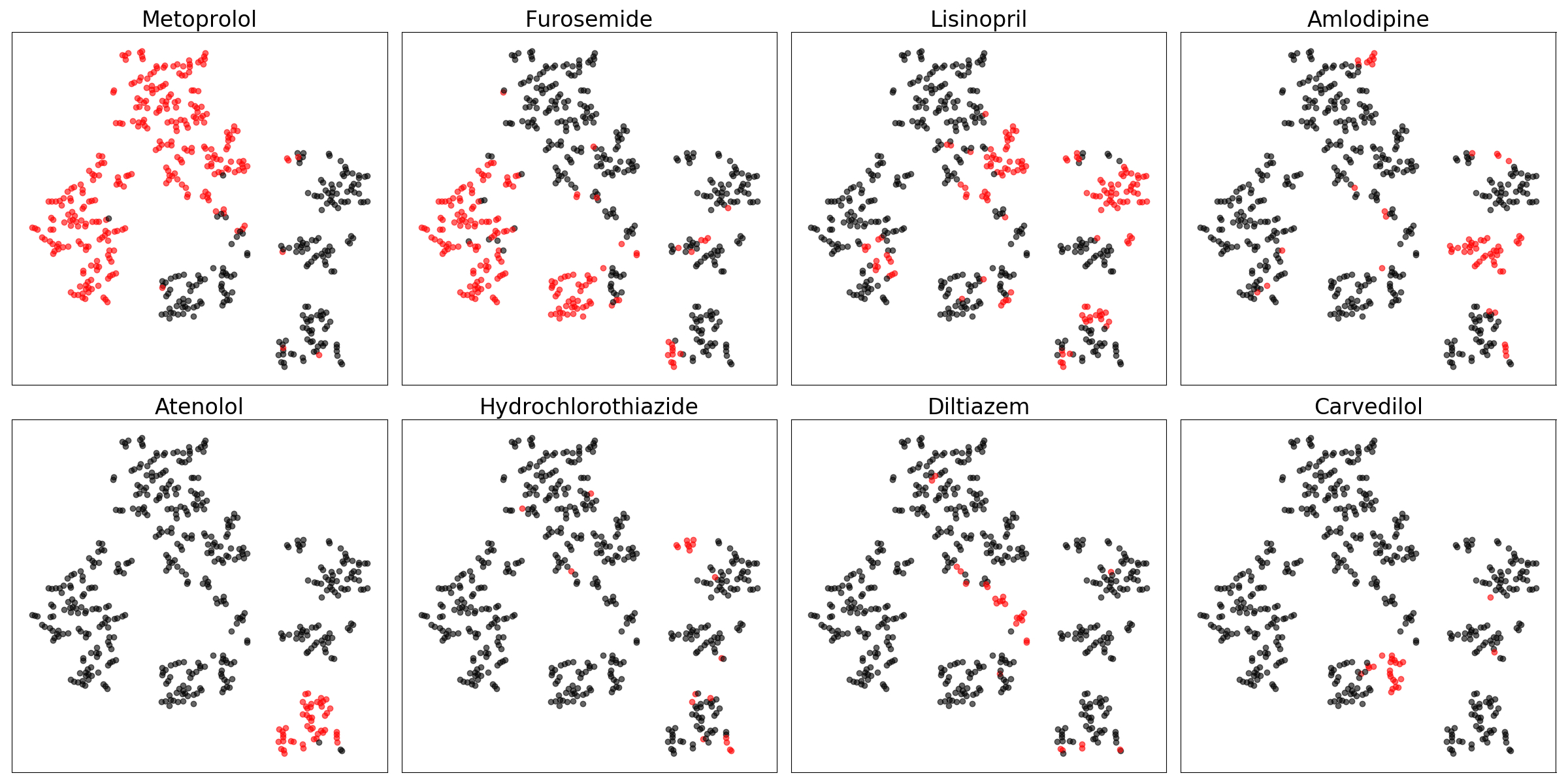}
\caption{T-SNE visualization of the latent vectors (denoted by dots) in the dense layer. Under each subplot titled with a medication, red dots denote the notes labeled with this medication and black dots denote those that are not.
%embedding space in hidden layer. Each subgraph corresponds to a certain medication class and only differs from the way of labelling: the point is labelled red if its groundtruth is positive for that class.
}
\label{fig:emb}
\end{figure}

One major reason that CNN outperforms other baseline models is that it uses a hierarchy of hidden layers to capture semantics in the notes at multiple granularities: word-level, n-gram level and note-level. We perform various visualizations to verify this. First, we examine whether the word embedding vectors are able to capture word-level semantics. For each word $w$, we compute the Euclidean distance between the embedding vector of $w$ and that of any other word, then retrieve the nearest neighboring word that has the smallest distances with $w$. Table \ref{tab:knn} shows the nearest neighbor of 20 exemplar words. Next, we visualize the filters in the convolutional layer and check whether they are able to capture semantics at the $n$-gram level. To visualize a filter, we apply it to all $n$-grams whose sizes match the window size of this filter and pick out the ones that yield the largest feature values. Table \ref{tab:phrase} shows the top $n$-grams for 4 filters of window size 3, and the other 4 filters of window size 4. Finally, we examine whether the dense layer vectors which represent the entire note are able to capture note-level semantics. We project these vectors from their original high-dimensional space into a two-dimensional space using the t-SNE~\cite{maaten2008visualizing} tool. As shown in Figure~\ref{fig:emb}, each dot represents a note. In each sub-plot titled with a medication name, red dots denote the notes that are associated with this medication and black dots denote the notes that are not associated with this medication.

Another source of improvement of the CNN model over the baselines is its ability in capturing the correlation among medications. To verify this, we compare the medication-correlations learned by our model and those computed from the training data. Let $\mb{A}=\text{cov}(\mb{y})$ (equation (\ref{eq:covy})) denote the covariance matrix learned by the model. The correlation between medication $i$ and $j$ is defined as
$$\text{corr}(i,j)=\frac{A_{ij}}{\sqrt{A_{ii}}\sqrt{A_{jj}}}.$$
On the other hand, we can measure the strength of association between the two medications using the point-wise mutual information (PMI) which is computed based on the co-occurrence statistics of the two medications:
$$\text{pmi}(i,j)=\log\frac{n(i,j)}{n(i)n(j)}.$$
$n(i,j)$, $n(i)$ and $n(j)$ are the number of admissions where $i$ and $j$ are used together, $i$ is used, $j$ is used, respectively. By definition, PMI quantifies the mutual dependence between two variables and can thus characterize the correlations between these variables. %The results are shown in Table~\ref{tab:cov}, where for each medication we list and rank the rest of the medications with respect to their covariance and PMI respectively.
 The results are shown in Table~\ref{tab:cov} where for each medication, we compute its correlation and PMI scores with all other medications (which are ranked in descending order of the corresponding scores).

% \usepackage{graphicx}
\begin{table}[t]
\centering
\begin{tabular}{ll}
\hline
\multicolumn{2}{c}{\textbf{Window Size 3}}                                          \\ \hline
\multicolumn{1}{c}{\textbf{Filter 1}}  & \multicolumn{1}{c}{\textbf{Filter 2}}      \\ \hline
edema varicosities glucose             & hypertension esrd dialysis                 \\
bilateral varicosities blood           & history esrd fistula                       \\
bilaterally varicosities patient       & syndrome esrd gastric                      \\
edema varicosities neuro               & black stools prior                         \\
superficial varicosities bilat         & procedures hemodialysis placed             \\ \hline
\multicolumn{1}{c}{\textbf{Filter 3}}  & \multicolumn{1}{c}{\textbf{Filter 4}}      \\ \hline
xanthalesma admission labs             & coronary artery saphenous                  \\
blood discharge labs                   & coronary artery bypass                     \\
blood mchc labs                        & valve disease replacement                  \\
blood notdone ctropnt                  & varicosities bilat post                    \\
ctropnt discharge labs                 & coronary disease pulse                     \\ \hline
\multicolumn{2}{c}{\textbf{Window Size 4}}                                          \\ \hline
\multicolumn{1}{c}{\textbf{Filter 5}}  & \multicolumn{1}{c}{\textbf{Filter 6}}      \\ \hline
short breath work symptoms             & combination dilated cardiomyopathy alcohol \\
distant breath sounds focal            & disease ischemic cardiomyopathy placement  \\
decreased breath sounds obese          & alcohol hypertensive cardiomyopathy etoh   \\
short breath male history              & congestive heart failure cardiac           \\
including phenylephrine paced function & diastolic heart failure fall               \\ \hline
\multicolumn{1}{c}{\textbf{Filter 7}}  & \multicolumn{1}{c}{\textbf{Filter 8}}      \\ \hline
coronary artery bypass grafting        & stopping congestive heart failure          \\
artery bypass grafting lima            & history congestive heart failure           \\
artery bypass grafts lima              & causing congestive heart failure           \\
artery bypass graft lima               & mild congestive heart failure              \\
coronary artery bypass grafts          & presented congestive heart failure         \\ \hline
\end{tabular}
\caption{Visualization of filters. For each filter, we show the top 5 $n$-grams that yield the largest feature values under this filter.}
\label{tab:phrase}
\end{table}

% \usepackage{graphicx}
\begin{table}[t]
\centering
\resizebox{\textwidth}{!}{%
\begin{tabular}{llll|llll|llll|llll}
\hline
\multicolumn{4}{c|}{Metoprolol}                     & \multicolumn{4}{c|}{Furosemide}                     & \multicolumn{4}{c|}{Lisinopril}                     & \multicolumn{4}{c}{Amlodipine}                     \\ \hline
\multicolumn{2}{c}{CORR} & \multicolumn{2}{c|}{PMI} & \multicolumn{2}{c}{CORR} & \multicolumn{2}{c|}{PMI} & \multicolumn{2}{c}{CORR} & \multicolumn{2}{c|}{PMI} & \multicolumn{2}{c}{CORR} & \multicolumn{2}{c}{PMI} \\
Furosemide    & 0.45     & Furosemide    & 0.02     & Carvedilol    & 0.54     & Carvedilol    & 0.12     & Atenolol      & 0.07     & Carvedilol    & 0.08     & Hctz          & 0.56     & Hctz          & 0.14    \\
Lisinopril    & -0.16    & Lisinopril    & -0.10    & Metoprolol    & 0.45     & Metoprolol    & 0.02     & Hctz          & 0.07     & Hctz          & 0.07     & Atenolol      & 0.37     & Carvedilol    & -0.14   \\
Amlodipine    & -0.27    & Diltiazem     & -0.21    & Amlodipine    & -0.26    & Diltiazem     & -0.17    & Carvedilol    & 0.06     & Atenolol      & 0.04     & Diltiazem     & 0.36     & Lisinopril    & -0.18   \\
Carvedilol    & -0.35    & Amlodipine    & -0.22    & Lisinopril    & -0.26    & Lisinopril    & -0.17    & Diltiazem     & -0.12    & Metoprolol    & -0.10    & Carvedilol    & -0.01    & Atenolol      & -0.18   \\
Hctz          & -0.35    & Hctz          & -0.24    & Diltiazem     & -0.28    & Atenolol      & -0.28    & Metoprolol    & -0.16    & Furosemide    & -0.17    & Furosemide    & -0.26    & Metoprolol    & -0.22   \\
Atenolol      & -0.45    & Atenolol      & -1.40    & Atenolol      & -0.32    & Amlodipine    & -0.28    & Furosemide    & -0.26    & Amlodipine    & -0.18    & Metoprolol    & -0.27    & Furosemide    & -0.28   \\
Diltiazem     & -0.47    & Carvedilol    & -1.64    & Hctz          & -0.50    & Hctz          & -0.72    & Amlodipine    & -0.41    & Diltiazem     & -0.26    & Lisinopril    & -0.41    & Diltiazem     & -0.62   \\ \hline
\multicolumn{4}{c|}{Atenolol}                       & \multicolumn{4}{c|}{Hctz}                           & \multicolumn{4}{c|}{Diltiazem}                      & \multicolumn{4}{c}{Carvedilol}                     \\ \hline
\multicolumn{2}{c}{CORR} & \multicolumn{2}{c|}{PMI} & \multicolumn{2}{c}{CORR} & \multicolumn{2}{c|}{PMI} & \multicolumn{2}{c}{CORR} & \multicolumn{2}{c|}{PMI} & \multicolumn{2}{c}{CORR} & \multicolumn{2}{c}{PMI} \\
Hctz          & 0.75     & Hctz          & 0.24     & Atenolol      & 0.75     & Atenolol      & 0.24     & Hctz          & 0.74     & Hctz          & 0.06     & Furosemide    & 0.54     & Furosemide    & 0.12    \\
Diltiazem     & 0.71     & Lisinopril    & 0.04     & Diltiazem     & 0.74     & Amlodipine    & 0.14     & Atenolol      & 0.71     & Furosemide    & -0.17    & Diltiazem     & 0.11     & Lisinopril    & 0.08    \\
Amlodipine    & 0.37     & Amlodipine    & -0.18    & Amlodipine    & 0.56     & Lisinopril    & 0.07     & Amlodipine    & 0.36     & Metoprolol    & -0.21    & Atenolol      & 0.06     & Amlodipine    & -0.14   \\
Lisinopril    & 0.07     & Furosemide    & -0.28    & Lisinopril    & 0.07     & Diltiazem     & 0.06     & Carvedilol    & 0.11     & Lisinopril    & -0.26    & Lisinopril    & 0.06     & Hctz          & -0.43   \\
Carvedilol    & 0.06     & Diltiazem     & -0.30    & Carvedilol    & -0.09    & Metoprolol    & -0.24    & Lisinopril    & -0.12    & Atenolol      & -0.30    & Amlodipine    & -0.01    & Diltiazem     & -0.75   \\
Furosemide    & -0.32    & Metoprolol    & -1.40    & Metoprolol    & -0.35    & Carvedilol    & -0.43    & Furosemide    & -0.28    & Amlodipine    & -0.62    & Hctz          & -0.09    & Metoprolol    & -1.64   \\
Metoprolol    & -0.45    & Carvedilol    & -1.93    & Furosemide    & -0.50    & Furosemide    & -0.72    & Metoprolol    & -0.47    & Carvedilol    & -0.75    & Metoprolol    & -0.35    & Atenolol      & -1.93   \\ \hline
\end{tabular}%
}
\caption{Medication correlation (CORR) scores learned by our model and point-wise mutual information (PMI) scores computed from training data.}
\label{tab:cov}
\end{table}

% \begin{figure}[tb]
% \hspace{0.15cm}
% \begin{subfigure}{.48\textwidth}
%     \centering
%     \includegraphics[width=\textwidth, trim={5cm 8cm 18cm 8cm}, clip]{figures/32520_trim.jpg}
%     \caption{}
%     \label{fig:doc_1}
% \end{subfigure}
% \hspace{0.05cm}
% \begin{subfigure}{.48\textwidth}
%     \centering    
% \includegraphics[width=\textwidth, trim={5cm 8cm 18cm 8cm}, clip]{figures/39111_trim.jpg}        
%     \caption{}
%     \label{fig:doc_2}
% \end{subfigure}

% \caption{Interpretation of CNN's predictions. In the two admission notes, words having strong association with the predicted medications are highlighted in red color. 
% %Visualization of how CNN weighs each word in the document when recommending medications: (a) metoprolol and lisinopril; and (b) lisinopril. Deeper red background color indicates higher word importance. 
% For better clarity, paragraphs containing no important words are not shown.
% }
% \label{fig:doc}
% \end{figure}

% ================================================
%                      Discussion
% ================================================
\section*{Discussion}
As shown in Table~\ref{tab:res}, on average, CNN outperforms the baseline models substantially. The reasons are two-fold. First, CNN has the ability to capture the semantics of admission notes at different granularities while the baselines methods lack such mechanisms. CNN learns the semantics of words, n-grams and entire notes using word-embedding layer, convolutional layer and dense layer respectively. 

As shown in Table~\ref{tab:knn}, for each query word, it has a strong clinical correlation with its nearest neighbor (NN) (which has the smallest Euclidean distance with the query word in the embedding space). For example, for ``artery '' and its NN ``troponin'', they are correlated in the following way: in patients with coronary artery disease, the level of troponin is elevated. ``aortic'' and ``mitral'' are correlated because the aortic valve and the mitral valve are the most commonly replaced valves. The correlation between ``angina'' and ``echocardiograms'' is: echocardiogram is used to identify the cause of angina. This demonstrates that the learned embedding vectors are able to well capture the semantics of words. 

%words that have small Euclidean distances in embedding vector space are strongly correlated in semantics. 

In the convolutional layer, each filter is able to capture a specific local semantic. As shown in Table~\ref{tab:phrase}, the top $n$-grams under a convolutional filter have strong semantic ties. For example, the top $3$-grams under Filter 1-4 are about varicosities, renal disease, lab tests and coronary disease respectively. The top $4$-grams under Filter 5-8 are about breathing problems, cardiomyopathy and heart failure respectively. Interestingly, grams that do not share any common words but possess related semantics are grouped under the same filter. For example, in Filter 2, the three grams ``hypertension esrd dialysis'', ``black stools prior'', ``procedures hemodialysis placed'' do not share any words in common, but their semantics are closely related: ``black stools'' is a common symptom of ``esrd'' (end-stage renal disease); ``hemodialysis'' is a common medical procedure of treating ``esrd''. Filter 8 demonstrates the necessity of using filters with different window sizes. While these $4$-grams are all about ``congestive heart failure'' (CHF), their difference in the first word reveals the subtlety in semantics: the 5 grams are about the ending, history, starting, severity and existence of CHF respectively. If we only use filters of window size 3, we can only capture one single semantic ``congestive heart failure''. But using filters of window size 4 enables us to capture more fine-grained semantics.

The note-level representation vector obtained in the dense layer is able to capture the global semantics of the entire note. As visualized in Figure~\ref{fig:emb}, for notes that are associated with the same discharge medication, their representation vectors are close to each other. For example, in the first subplot, the notes are clearly clustered into two groups, which correspond to ``with metoprolol" and ``without metoprolol'' respectively. The similar phenomenon is seen in other plots as well. This indicates that these concatenation vectors are very informative for predicting discharge medications. In addition, from these plots, we can identify a few clinical insights. First, from the subplots titled with ``metoprolol'' and ``furosemide'', we can see the red points have substantial overlap, which indicates that these two medications tend to be used together. Second, from the plots titled with ``amlodipine'' and ``atenolol'', we can see that these two medications have minimal overlap, indicating that they are seldom used together.

The other major reason that the CNN model outperforms the baselines is because it is able to capture the correlation among medications. As can be seen from Table~\ref{tab:cov}, the rankings based on correlation are very consistent with those based on PMI. For example, for all medications except lisinopril, the most correlated medication found according to correlation is the same as that found according to PMI. This indicates that the CNN model can effectively capture the correlations among medications.

Comparing the F1 scores on individual medications, we can see that CNN achieves larger improvements on less frequent medications such as atenolol and diltiazem. On frequent mediations such as metoprolol and furosemide, CNN is on par with the baseline methods. As a consequence, CNN achieves more improvement on macro average F1 which does not take into account medication frequency. For the micro average F1 where the individual F1 scores are weighted using the frequency of medications, CNN's improvement over the baselines is less significant. We conjecture that CNN's better performance on infrequent medications lies in its ability to capture medication correlations. The prediction of a medication depends on two factors: (i) how strongly the medication is relevant to the input admission note; (ii) how strongly this medication is correlated with other medications. For infrequent medications where a high-quality model measuring (i) is difficult to train due to the lack of training examples, (ii) becomes very important for making the prediction correct. CNN has the machinery to effectively capture medication correlations, hence works better on infrequent medications.  

MLP achieves the worst F1 on 5 individual medications and the worse micro average F1. The reason is it only uses admission medications as inputs while the rest of methods also leverage other admission information such as past medical history and chief complaint. This indicates that the discharge medications are remarkably different from the admission medications and other types of information must be effectively leveraged to make accurate predictions. For all models, better F1 scores are achieved on medications with larger frequencies. This is reasonable since larger training data typically leads to better predictive performance in machine learning. 

\section*{Limitations}

While CNN yields better performance than other strong baseline models, we discuss a few of its limitations. First of all, the gap between CNN's performances on frequent and infrequent medication classes is still large, though it has been improved compared with the baselines. For example, the F1 scores achieved by CNN on two most frequent medications are 0.79 and 0.70 respectively, which are significantly better than those on infrequent medications such as atenolol and hctz. In future work, we plan to come up with methods to bridge this gap. 

Another limitation of our method is that it is purely data-driven and does not incorporate human knowledge. In clinical practice, physicians refer to guidelines made by professional associations to prescribe medications. Such guidelines can be incorporated into the CNN model to further improve the prediction accuracy, which we plan to investigate in future.

In our current approach, only medical information is leveraged to predict the discharge medications. Non-clinical factors, such as insurance type, cost of medications, influence the prescription of medications as well, which should be incorporated in the prediction model. 

The notes in MIMIC-III are highly noisy and our current pre-processing steps do not completely deal with these noises. For example, the heading strings (Table~\ref{tab:syns}) used to identify different types of admission information are not exhaustive, which results in a lot of missing information. For the next step, we are going to manually deal with these corner cases and include them into the training set.

% ================================================
%                     Conclusion
% ================================================
\section*{Conclusion}

We find that it is possible to accurately predict discharge medications only using the information available at admission time. Such predictions can provide valuable information for physicians to perform treatment planning. On 8 medications, the CNN model achieves a (micro-averaged) precision of 0.63 for a recall of 0.70.
%, suggesting that deep learning approaches have the potential to accurately predict discharge medications. 

The CNN model outperforms the best baseline model for over 20\% in terms of the macro-averaged F1-score. The performance gain is attributed to CNN's two abilities that are not owned by the baseline methods. First, CNN is able to learn semantic representations of texts. We perform detailed visualizations of the individual model-components, including word embeddings, convolutional filters and dense layers. These visualizations show that the CNN model is able to distill semantics with different granularities from the raw texts. Second, CNN is able to capture the correlations among medications thanks to the mechanism of sharing common latent factors. This ability is the other major reason that CNN performs better than the baselines, especially on infrequent medications, where the captured correlation remedies the lack of training examples.

Although our solution is motivated from a specific task, it is potentially generic for other clinical predictive tasks. For example, by switching target label from medications to diseases, CNN can be used to aid comorbidity diagnosis.

%All the representations are learned in an end-to-end and task-specific way. Guided by the ground-truth discharge medications, the representations are learned to be particularly suitable for predicting discharge medications. In the baseline methods, the representations of admission notes are fixed and not adjustable. These representations are task-agnostic and not particularly informative for our task. 

% ================================================
% 			      	     ref
% ================================================
\bibliography{sample}

% ================================================
% 			      Acknowledgements
% ================================================
% \section*{Acknowledgements (not compulsory)}

% Acknowledgements should be brief, and should not include thanks to anonymous referees and editors, or effusive comments. Grant or contribution numbers may be acknowledged.
% \yy{TODO}

\section*{Author contributions statement}

% Must include all authors, identified by initials, for example:
% A.A. conceived the experiment(s),  A.A. and B.A. conducted the experiment(s), C.A. and D.A. analysed the results.  All authors reviewed the manuscript. 

P.X., C.C. and E.X. conceived the study. Y.Y., P.X. and E.X. designed the study. Y.Y. and P.X. wrote the main manuscript text. Y.Y. prepared all figures and tables. X.G. and C.C. helped writing the medical part of the manuscript. C.L. and H.Z. helped with model implementation and data preparation.

\section*{Additional Information}

\subsection*{Competing Interests}

The author(s) declare no competing financial interests.

\subsection*{Supplementary Information}

Table \ref{tab:std} shows the standard deviations of medication-wise precision, recall and F1-score of all models in 5 runs.

% \usepackage{multirow}
% \usepackage{graphicx}
\begin{table}[h]
\centering
\resizebox{\textwidth}{!}{%
\begin{tabular}{c|ccc|ccc|ccc|ccc|ccc}
\hline
\multirow{2}{*}{Medication} & \multicolumn{3}{c|}{CNN} & \multicolumn{3}{c|}{MLP} & \multicolumn{3}{c|}{SVM} & \multicolumn{3}{c|}{RF} & \multicolumn{3}{c}{LR} \\ \cline{2-16} 
                            & P      & R      & F      & P      & R      & F      & P      & R      & F      & P      & R      & F     & P      & R     & F     \\ \hline
Metoprolol                  & 0.02   & 0.03   & 0.01   & 0.02   & 0.04   & 0.01   & 0.01   & 0.01   & 0.00   & 0.00   & 0.01   & 0.00  & 0.01   & 0.01  & 0.01  \\
Furosemide                  & 0.01   & 0.03   & 0.01   & 0.10   & 0.18   & 0.05   & 0.01   & 0.02   & 0.01   & 0.01   & 0.01   & 0.01  & 0.01   & 0.02  & 0.02  \\
Lisinopril                  & 0.04   & 0.03   & 0.02   & 0.04   & 0.13   & 0.10   & 0.02   & 0.02   & 0.01   & 0.01   & 0.01   & 0.01  & 0.02   & 0.01  & 0.01  \\
Amlodipine                  & 0.09   & 0.13   & 0.09   & 0.23   & 0.16   & 0.19   & 0.02   & 0.03   & 0.03   & 0.04   & 0.03   & 0.03  & 0.06   & 0.02  & 0.03  \\
Atenolol                    & 0.10   & 0.14   & 0.14   & 0.00   & 0.00   & 0.00   & 0.03   & 0.01   & 0.00   & 0.06   & 0.01   & 0.02  & 0.08   & 0.01  & 0.01  \\
Hctz         & 0.21   & 0.20   & 0.21   & 0.00   & 0.00   & 0.00   & 0.04   & 0.02   & 0.02   & 0.04   & 0.01   & 0.02  & 0.23   & 0.01  & 0.02  \\
Diltiazem                   & 0.05   & 0.06   & 0.06   & 0.00   & 0.00   & 0.00   & 0.09   & 0.05   & 0.06   & 0.18   & 0.05   & 0.07  & 0.14   & 0.02  & 0.03  \\
Carvedilol                  & 0.11   & 0.02   & 0.06   & 0.00   & 0.00   & 0.00   & 0.08   & 0.04   & 0.05   & 0.06   & 0.01   & 0.02  & 0.13   & 0.03  & 0.05  \\ \hline
\end{tabular}%
}
\caption{Standard deviations of medication-wise precision (P), recall (R) and F1-score (F) for CNN and 4 baseline models averaged over 5 runs.}
\label{tab:std}
\end{table}

\iffalse
\begin{figure}[ht]
\centering
\includegraphics[width=\linewidth]{stream}
\caption{Legend (350 words max). Example legend text.}
\label{fig:stream}
\end{figure}

\begin{table}[ht]
\centering
\begin{tabular}{|l|l|l|}
\hline
Condition & n & p \\
\hline
A & 5 & 0.1 \\
\hline
B & 10 & 0.01 \\
\hline
\end{tabular}
\caption{\label{tab:example}Legend (350 words max). Example legend text.}
\end{table}

Figures and tables can be referenced in LaTeX using the ref command, e.g. Figure \ref{fig:stream} and Table \ref{tab:example}.
\fi
\end{document}